\documentclass[11pt]{article}
\usepackage[preprint]{acl}
\usepackage{times}
\usepackage{latexsym}
\usepackage[T1]{fontenc}
\usepackage[utf8]{inputenc}
\usepackage{microtype}
\usepackage{inconsolata}
\usepackage{amsmath,amssymb,booktabs,array,graphicx,tikz}
\newcommand{\undet}{\texttt{UNDETERMINED}}
\newcommand{\ind}{\mathbb{I}}
\newcommand{\suff}{\mathrm{S}}
\newcommand{\insuff}{\mathrm{I}}
\newcommand{\unk}{\mathrm{U}}
\title{Correct Answers from Insufficient Histories:\\A Controlled Study of Agent Memory Evaluation}
\author{%
\textbf{Quan Xue, Zerui Yang, Chaorui Zhang, Haolin Ruan,}\\
\textbf{Xueyan Niu, Xuanyu Zheng, Xiyuan Wang,}\\
\textbf{Weixi Zhang, Wei Han, Bo Bai}\\
\textnormal{Huawei Technologies Ltd.}\\
\textnormal{Correspondence: Quan Xue (\texttt{csxuequan@connect.hku.hk});}\\
\textnormal{Chaorui Zhang (\texttt{zhang.chaorui@huawei.com})}
}
\date{}
\hypersetup{
  pdftitle={Correct Answers from Insufficient Histories: A Controlled Study of Agent Memory Evaluation},
  pdfauthor={Quan Xue, Zerui Yang, Chaorui Zhang, Haolin Ruan, Xueyan Niu, Xuanyu Zheng, Xiyuan Wang, Weixi Zhang, Wei Han, Bo Bai}
}

\begin{document}
\maketitle

\begin{abstract}
An answer can match the reference for a complete agent history even when the reader's visible history does not determine it. We study this distinction using executable device logs. Our protocol separates reference agreement, certified evidence sufficiency, and context-appropriate behavior. Insufficiency requires two legal histories yielding byte-identical visible contexts but different answers; public chronology and query cutoffs certify sufficient prefixes. On 80 confirmation histories, Qwen3-32B matches 87 hidden references among 203 insufficient raw-prefix instances, with only seven appropriate abstentions. A fixed-input DeepSeek-V4-Flash replication matches 85 and abstains on 21 of the same 203 instances. Each reader makes 1,600 requests. Compact CSV's estimated appropriateness effects remain inconclusive: $+3.16$ percentage points, clustered 95\% interval $[0.00,6.57]$, for Qwen and $+2.46$, $[-1.84,6.94]$, for DeepSeek. A separate held-out judge study achieves 99.38\% correctness but retains support errors and truncations; a deterministic baseline solves all constructed cases. These results support an evaluation diagnosis across two hosted reader configurations, not gains for a general memory system or natural-language evidence assessment.
\end{abstract}

\section{Introduction}
Memory systems expose a selected or compressed part of an interaction history to a language model. Their evaluation often ends by comparing the model's answer with a reference derived from the complete history. This comparison answers a useful question: did the system recover the expected value? It leaves another question unresolved: did the information actually supplied to the reader determine that value? A model can agree with the hidden reference after a decisive tool outcome has been removed. Conversely, a truncated history can already determine an answer when public temporal constraints exclude any relevant unseen event.

Consider a device whose visible log contains one accepted route to channel A. If the original history may continue, the visible text does not determine the final channel: a hidden accepted route to B would change it. The answer A can nevertheless match the reference in the observed original history. For a query asking about the state before a cutoff, a chronological prefix containing a complete record after that cutoff can be sufficient even if the full log is absent. The relevant distinction is therefore neither answer correctness alone nor whether the context was truncated.

Prior work already separates context sufficiency from model success in retrieval-augmented generation \citep{joren2025sufficient}, and answerability is central to abstention benchmarks \citep{rajpurkar2018know}. We do not introduce this distinction. We investigate a narrower measurement problem in history-based question answering: how to certify it when evidence contains repeated reports, unsuccessful actions, temporal scopes, and state transitions. These operations arise in agent trajectories \citep{zhao2026ama}, while conversational memory evaluations test related temporal and update abilities \citep{maharana2024locomo,wu2025longmemeval}.

We use a small executable environment to make the evidence boundary inspectable. A reference answer is checked against environment snapshots, Python replay, and a separate SQL implementation. A concrete answer counts as appropriate only when the visible context uniquely determines the answer under the public rules. Otherwise the prescribed response is \undet. We conservatively certify evidence status and retain unclassified cases rather than infer sufficiency from a failed counterexample search.

Our experiments address three questions. \textbf{RQ1:} How often does reference agreement occur under certified insufficiency? \textbf{RQ2:} Does compact formatting improve context-appropriate behavior at a fixed context budget when some partial histories are sufficient? \textbf{RQ3:} Can a model judge reliably distinguish reference agreement from context appropriateness, and does it improve on simple executable baselines?

The diagnosis appears in two hosted reader configurations. Qwen abstains on only 1/160 insufficient outcome-masked inputs while matching 72 hidden references. On chronological raw prefixes, Qwen and DeepSeek give concrete answers on 196/203 and 182/203 insufficient instances, respectively. Neither reader's budgeted CSV appropriateness interval is strictly positive; the positive Qwen full-information contrast is also not replicated by DeepSeek. Separately, the judge's 99.38\% score coexists with support errors, and a deterministic baseline matches the hybrid's observed perfect score.

Our contribution is an executable, auditable evaluation protocol and a controlled empirical diagnosis, rather than a new retrieval or memory architecture. Specifically, we provide (i) explicit task-relative evidence certificates, including same-context counterexamples; (ii) reader comparisons that report sufficient, insufficient, and unclassified inputs separately; and (iii) held-out constructed controls for evaluator reliability, with strong program baselines and a history-level analysis. All claims below are restricted to the declared grammar, generator, and hosted model configurations.

\section{Related Work}
\paragraph{Memory and context use.}
LoCoMo evaluates long-term conversational memory through question answering, summarization, and dialogue generation \citep{maharana2024locomo}. LongMemEval tests extraction, reasoning across sessions, temporal reasoning, updates, and abstention, and separates indexing, retrieval, and reading \citep{wu2025longmemeval}. AMA-Bench extends evaluation to agent--environment trajectories \citep{zhao2026ama}. We study a controlled reading and evaluation component; our synthetic logs are not a replacement for these benchmarks. Work on long-context use also shows that supplying information does not guarantee a model will use it successfully \citep{liu2024lost}. Our full-information format controls examine a related failure mode without claiming a test of long-horizon scaling.

\paragraph{Sufficiency, answerability, and grounding.}
\citet{joren2025sufficient} distinguish sufficient from insufficient retrieved context and analyze selective generation. SQuAD 2.0 requires abstention when a passage does not support an answer \citep{rajpurkar2018know}. KILT evaluates knowledge-intensive tasks with provenance \citep{petroni2021kilt}; ALCE assesses generated citations \citep{gao2023alce}; FActScore evaluates supported atomic facts \citep{min2023factscore}. Our setting differs in what can be certified: fixed tool semantics permit an explicit pair of compatible histories with different query answers. A quote, a matching value, or a relevant record alone does not certify exhaustive counts or final states. We add no claim of general natural-language entailment verification.

\paragraph{Automatic evaluation.}
LLM-based judges offer scalable evaluation but exhibit biases and reasoning limitations \citep{zheng2023judging}. RAGAs separates aspects of retrieval and generation quality \citep{es2024ragas}. We likewise separate axes, then test the evaluator on constructed cases with executable labels. This differs from estimating human preference agreement: our controls concern a finite answer grammar and a specified interpretation of conflicting assertions. We compare against rules that already solve that grammar, so improvements over a weaker model configuration do not establish an advantage over the strongest simple baseline.

\paragraph{Concurrent agent-memory studies.}
\citet{anonymous2026tracing} isolate query and packing failures in TRACE through planner-input and fixed-input selection interventions. Their question is which pipeline component explains an answer improvement; ours is whether the supplied history determines the answer, established through same-context witnesses and public sufficiency rules. \citet{anonymous2026rdp} formalize query-conditioned rate--distortion--perception trade-offs, prove a coverage-necessity lemma, and evaluate a selector on LoCoMo and LongMemEval. Our target is answer identifiability under event semantics, distinct from distributional resemblance or selector performance. \citet{anonymous2026aggregation} introduce source-preserving Trajectory Memory and TallyMem's event specifications and deterministic aggregation, evaluated on MINTEval and AMA-Bench. Both studies use executable event processing: TallyMem computes answers within a memory system, whereas our executors verify references and evidence certificates on synthetic device histories with duplicates, rejected actions, and query cutoffs. These concurrent studies motivate complementary questions about attribution, compaction, and aggregation; our contribution concerns evaluation validity and appropriate answering or abstention.

\section{Task and Evidence Certificates}
\label{sec:task}
\subsection{Histories and Public Semantics}
A history consists of reports with eight fields: report identifier, report time, event identifier, event time, device, action, argument, and outcome. Reports can duplicate an event. Duplicate reports share all event fields and count once. Event time determines execution order; report time does not. Each device starts in \texttt{UNSET}. An \texttt{ACCEPTED route} sets its channel to the argument; an \texttt{ACCEPTED disable} sets it to \texttt{DISABLED}; a \texttt{REJECTED} action has no effect.

The four query operators are the number of distinct accepted routes, first accepted route channel, last accepted route channel, and device state. First and last return \texttt{NONE} when there is no accepted route. Count returns an integer. State returns a channel, \texttt{DISABLED}, or \texttt{UNSET}. In the chronological study, queries specify an inclusive cutoff $T$ and use only events with time $t\leq T$. Queries follow a fixed English grammar. These are operation and evidence tests, not open-ended language-understanding tasks.

\subsection{Task-Relative Sufficiency}
Let $H$ denote a complete legal history, $q$ a query, $P$ a context packer, and $C=P(H)$ the visible context. Let $\mathcal{G}$ contain only the guarantees disclosed to the reader, including the task semantics and the permitted missing-information mechanism. Define the compatible histories as
\begin{equation}
\mathcal{W}(C;P,\mathcal{G})=\{H':H'\models\mathcal{G},\ P(H')=C\}.
\end{equation}
For outcome masking, compatibility instead fixes the visible fields and allows the declared missing outcomes to vary. If $f(H,q)$ is the executable answer, the context is sufficient exactly when
\begin{equation}
\left|\{f(H',q):H'\in\mathcal{W}(C;P,\mathcal{G})\}\right|=1.
\label{eq:suff}
\end{equation}
The original history belongs to this set. Sufficiency is relative to $\mathcal{G}$, not to the generator's private event count or probability distribution. For example, a reader cannot infer that no unseen event exists merely because the experiment happens to generate a fixed number of events. This distinction prevents a hidden construction choice from becoming an unstated proof premise.

We do not exhaustively solve Equation~\ref{eq:suff} for every partial string. Each input receives one of three conservative labels: \textbf{sufficient}, with a public proof; \textbf{insufficient}, with a verified counterexample; or \textbf{unclassified}, when neither certificate is found. Full-source and partial-source sufficiency are recorded separately. Unclassified does not mean insufficient.

\begin{figure}[t]
\centering
\fbox{\begin{minipage}{0.91\columnwidth}
\small
\textbf{Visible chronological prefix}\par
\texttt{t=10: route(A), ACCEPTED}\par
\textit{No complete-log marker; query cutoff $T=20$.}\par\smallskip
\textbf{Two compatible worlds}\par
$H_1$: the log ends here; state at $T$ is A.\par
$H_2$: an unseen route to B occurs at $t=20$; state is B.\par
Both must reproduce the same packed bytes.\par\smallskip
\textbf{Query-dependent boundary}\par
The first route is already A in both worlds. State, count, and last route can still differ. A fully visible record at $t>20$ would close the cutoff scope.
\end{minipage}}
\caption{Illustrative evidence certificates. The example explains the public rules; it is not a sampled model response. A witness must be legal and pass actual repacking, not merely describe a possible hidden event.}
\label{fig:worlds}
\end{figure}

\subsection{Certificates and Their Scope}
\paragraph{Insufficiency.}
We store the original and an alternative legal history, verify different answers with SQL and Python, and run the identical tokenizer and packer on both. The resulting contexts must be byte-identical. Candidate alternatives flip a hidden outcome, change a hidden route argument, or add a legal suffix event at the cutoff. Changes to a duplicated event are synchronized across its reports. Chronology, distinct event times, and duplicate consistency are checked again. This is a constructive certificate that a concrete answer is not forced by the visible information.

\paragraph{Sufficiency.}
Three public conditions suffice in the chronological task. (1) A complete-log marker certifies the entire source. (2) A complete visible record strictly after $T$, together with a contiguous chronological prefix, certifies that all events at or before $T$ are present. (3) For the first-route query, the first visible accepted route at or before $T$ already fixes the answer, even if the cutoff has not been crossed. The value is computed from complete records using the public semantics. Partial trailing fields are preserved but are not promoted to complete records.

The second condition relies on ordering by event time and on the prefix starting at the beginning of the log. It would not hold for arbitrary retrieved chunks. The third holds for first route but not generally for count, last route, or state. These conditions are sufficient rather than necessary; a partial field can sometimes establish more than our parser certifies. We retain those cases as unclassified. Figure~\ref{fig:worlds} illustrates the query dependence.

\subsection{Separate Outcome Measures}
Let $a$ be the reader's parsed answer, $y=f(H,q)$ the complete-history reference, and $z\in\{\suff,\insuff,\unk\}$ the evidence label. Reference agreement is $R=\ind[a=y]$, even on insufficient contexts. For certified inputs, context appropriateness is
\begin{equation}
A=\begin{cases}
\ind[a=y],&z=\suff,\\
\ind[a=\undet],&z=\insuff.
\end{cases}
\label{eq:appropriate}
\end{equation}
Malformed outputs fail the output contract and do not receive credit. We report the certified denominator, reference hits on insufficient inputs, abstention on insufficient inputs, and incorrect abstention on sufficient inputs. For $c$ appropriate answers, $u$ unclassified inputs, and $n$ total inputs, the all-input appropriateness interval is $[c/n,(c+u)/n]$. This is an identification bound, not a confidence interval.

Appropriateness implements the declared contract to answer only when the value is determined. It is not a utility-optimal policy for every application: an application may prefer a probabilistic guess to abstention. Such a policy needs a different objective and a stated cost of error. The reference score remains useful, but answers a different measurement question.

\section{Experimental Design}
\label{sec:design}
\subsection{Studies and Exposure}
Table~\ref{tab:studies} distinguishes development, confirmation, and replication. Masking and unordered packing use 40 exposed histories; the judge uses 120 reserved histories from that generator. Qwen's chronological study uses a new generator, 16 development histories, and 80 confirmation histories from an independent random stream. DeepSeek reuses those 80 histories after Qwen's results are observed. We do not count replication as new histories or external-domain confirmation.

\begin{table*}[t]
\centering\small
\begin{tabular}{@{}>{\raggedright\arraybackslash}p{3.35cm}>{\raggedright\arraybackslash}p{3.05cm}r >{\raggedright\arraybackslash}p{5.25cm}@{}}
\toprule
Study & Histories / status & Model requests & Interpretation \\
\midrule
Outcome masking & 40 development & 480 & Equal-length evidence intervention; four queries per history \\
Unordered packing & Same 40 development & 2,080 & Budget and format exploration, including supplement; shared requests \\
Judge confirmation & 120 reserved & 2,400 & Constructed candidates; no new reader answers \\
Chronological Qwen & 16 development + 80 confirmation & 1,920 & Five contexts and four queries; 1,600 confirmation requests \\
DeepSeek replication & Same 80, Qwen-exposed & 1,600 & Identical visible inputs; second reader configuration \\
\bottomrule
\end{tabular}
\caption{Distinct experimental roles. Request counts are successful unique requests, except that the judge count includes explicitly independent stress-test repetitions. They are not sample sizes for statistical inference. The unordered packing count includes its later 1,280-token supplement.}
\label{tab:studies}
\end{table*}

\subsection{Reader and Packing Conditions}
We use hosted \texttt{Qwen/Qwen3-32B} \citep{qwen2025qwen3} and \texttt{deepseek-ai/DeepSeek-V4-Flash} \citep{deepseek2026v4} at temperature 0, \texttt{reasoning\_effort=none}, with a 64-token output limit. Prompts state the semantics, request \undet\ when evidence does not determine an answer, and explain chronological sufficiency. This is an instructed diagnostic. Successful contents, including errors, are retained once per unique request.

Raw logs serialize the eight fields into readable reports. CSV serializes the same field values with one header and one line per report; round-trip checks ensure that the complete CSV is lossless. Raw-prefix and CSV-prefix policies truncate to a context-token budget and may cut a record. Whole-record raw prefixes stop before the next record would exceed that budget. All full-source conditions include the same complete-log marker. We budget only the context using the frozen local Qwen tokenizer; instructions and questions are additional input. Consequently, equal context budgets do not mean equal billed API prompt counts. Whole-record prefixes may use less than the maximum budget.

The unordered development study tests 256, 512, 1,024, and 2,048 context tokens, plus full-source controls. A supplementary 1,280-token budget is chosen from input lengths before its new reader calls. The chronological study freezes a 768-token context budget and full raw/full CSV controls. Each of its histories contains 24 distinct events, four extra duplicate reports, and three devices. Reports are ordered by event time, distinct events have distinct times, and queries use an inclusive cutoff. The event count and generator distribution are not disclosed to the reader or used in certificates.

DeepSeek's replication changes only the model field of Qwen's frozen request payloads, preserving messages, seed, and strict answer schema. Visible UTF-8 bytes are identical; 768 tokens refer to the original Qwen tokenizer, not an equal DeepSeek-native budget. Inputs, analysis, and an eight-request technical pilot are frozen before DeepSeek calls. The pilot checks service and output contracts, not accuracy, and remains in the 1,600 requests.

\subsection{Validation and Statistical Analysis}
Gold answers are checked against the generator's environment snapshots, a Python replay, and SQL deduplication and temporal queries. This is implementation agreement within declared semantics, not human validation of unrestricted language. Certificate audits check public premises and exact repacking. Deterministic program baselines accept only the actual context and question, not private labels.

We resample whole histories with replacement, retaining all questions and paired conditions, for 10,000 bootstrap replicates. Reported intervals are percentile 95\% intervals. In the chronological primary contrast, CSV minus raw appropriateness is computed on questions certified in \emph{both} arms. Resampling then recomputes each numerator/denominator ratio. Comparing marginal rates with different certification coverage would target a different population. The positive-effect criterion, fixed before model calls, requires the lower interval endpoint to be strictly above zero. Secondary format and subgroup results are descriptive; we do not claim a family-wise discovery guarantee.

Qwen's chronological development gate checks executable truth, witnesses, sufficient and insufficient counts, at most 10\% unclassified inputs per prefix arm, service contracts, and at least 50\% full-source agreement. All checks pass without design changes before confirmation is generated. DeepSeek's separately frozen replication criterion requires the insufficient raw-prefix concrete-answer rate's lower interval endpoint to exceed 50\% and the unsupported-reference-hit rate's to exceed zero. These are operational diagnostic thresholds. Logs, manifests, and network-disabled replay are retained (Appendix~\ref{app:reproduce}).

\section{Reader Results}
\label{sec:reader}
\subsection{RQ1: Agreement without Sufficient Evidence}
\paragraph{Controlled outcome masking.}
For each of 160 development queries, we compare a complete log, a decisive outcome replaced with \texttt{WITHHELD}, and an irrelevant outcome replaced with the same marker. All duplicate reports of the selected event are masked together. Two possible outcome assignments change the answer in the decisive arm; the irrelevant arm preserves a unique answer. Neutral padding equalizes local context-token counts.

Qwen is appropriate on 111/160 complete inputs (69.4\%; clustered 95\% CI $[62.5,76.2]$) and 112/160 irrelevant-mask inputs (70.0\%; $[63.1,76.9]$). The complete-minus-irrelevant difference is $-0.6$ percentage points (pp), $[-3.8,2.5]$. On decisive masking, only 1/160 responses abstains appropriately (0.6\%; $[0.0,1.9]$), yet 72/160 match the hidden complete-history reference. Thus reference agreement can remain substantial after evidence becomes provably insufficient.

This intervention is an oracle diagnostic with an explicit missing-outcome marker. It does not estimate the frequency of natural retrieval failures. Moreover, 47/160 decisive events have two reports while the chosen irrelevant event has one. Equal token length does not isolate every perturbation property; we do not interpret the decisive-versus-irrelevant gap as a clean causal effect of semantic relevance alone.

\paragraph{Unordered prefix development.}
In Qwen's unordered study, at 1,024 tokens raw prefixes match 96/160 references and CSV prefixes match 106/160; neither arm abstains. Their reference difference is $+6.25$ pp, with CI $[0.00,13.13]$. An initial certificate search based on changing one outcome certifies 1,036 non-full inputs as insufficient and leaves 404 unclassified across the primary budget grid. A later, explicitly post-collection audit adds legal hidden events because the public rules do not bound original log length. It certifies all 1,440 non-full primary inputs as insufficient. We preserve the original labels and separate this broader witness search from the original analysis.

Under that supplementary audit, all 96 and 106 reference hits at 1,024 tokens occur on insufficient contexts. The apparent reference improvement is therefore not an increase in answers supported by the supplied context. At the input-selected 1,280-token supplement, CSV happens to include all sources while raw remains incomplete; CSV obtains 126/160 appropriate answers and raw obtains 0/160. Completeness and the end marker vary together, so that comparison cannot isolate a formatting mechanism.

\subsection{RQ2: Chronological Prefix Confirmation}
The unordered task admits a degenerate policy: abstain whenever the complete-log marker is missing. The chronological study deliberately includes partial but sufficient contexts, so this policy can no longer solve the task. Across its three partial arms, confirmation has 327 sufficient, 585 insufficient, and 48 unclassified input--query conditions. The full-source arms add 640 sufficient conditions. Counts here include paired conditions, not independent questions.

\begin{table*}[t]
\centering\small
\setlength{\tabcolsep}{3pt}
\begin{tabular}{@{}lrrrrrrr@{}}
\toprule
 & \multicolumn{2}{c}{Qwen3-32B} & \multicolumn{2}{c}{DeepSeek-V4-Flash} & \multicolumn{3}{c}{Shared evidence counts} \\
\cmidrule(lr){2-3}\cmidrule(lr){4-5}\cmidrule(l){6-8}
Context & Reference & Appropriate & Reference & Appropriate & Partial S & I & U \\
\midrule
Raw prefix & 181/320 & 90/307 & 172/320 & 98/307 & 104 & 203 & 13 \\
Whole-record raw & 189/320 & 89/307 & 172/320 & 94/307 & 104 & 203 & 13 \\
CSV prefix & 195/320 & 108/298 & 167/320 & 111/298 & 119 & 179 & 22 \\
Full raw & 166/320 & 166/320 & 198/320 & 198/320 & --- & 0 & 0 \\
Full CSV & 188/320 & 188/320 & 196/320 & 196/320 & --- & 0 & 0 \\
\bottomrule
\end{tabular}
\caption{Qwen confirmation and fixed-input DeepSeek replication: the same 80 histories, four queries, five contexts, 1,600 requests per reader. S, I, U denote sufficient, insufficient, and unclassified; full-source cases are sufficient. Appropriate denominators exclude U. Raw and CSV prefixes use 768 Qwen context tokens; whole-record raw averages 728.05. Visible texts are identical across readers.}
\label{tab:reader}
\end{table*}

Table~\ref{tab:reader} reports both readers. Qwen's raw and CSV reference rates are 56.6\% and 60.9\%; certified appropriateness is 29.3\% and 36.2\% on different denominators. The paired contrast uses 285 questions certified in both arms, excluding 35 unclassified on at least one side. CSV minus raw appropriateness is $+3.16$ pp, CI $[0.00,6.57]$, for Qwen and $+2.46$ pp, $[-1.84,6.94]$, for DeepSeek. Neither lower endpoint is strictly positive. These inconclusive results do not establish equivalence.

Certification coverage itself changes with packing. Of 320 paired questions, 104 remain partial sufficient, 179 remain insufficient, two move from insufficient to sufficient, 13 move from unclassified to sufficient, and 22 move from insufficient to unclassified. CSV exposes more complete records, but a different partial tail can also reduce certification coverage. The primary certified intersection omits these unclassified cases by design. All-input appropriateness bounds are $[28.1,32.2]$\% for raw, $[27.8,31.9]$\% for whole-record raw, and $[33.8,40.6]$\% for CSV. These bounds describe each arm; their endpoints are not a paired uncertainty interval.

\begin{table}[t]
\centering\small
\begin{tabular}{@{}lrr@{}}
\toprule
Certified raw-prefix subset & Qwen & DeepSeek \\
\midrule
Sufficient: correct & 83/104 & 77/104 \\
Sufficient: abstains & 4/104 & 23/104 \\
Insufficient: concrete & 196/203 & 182/203 \\
Insufficient: abstains & 7/203 & 21/203 \\
Insufficient: ref. hit & 87/203 & 85/203 \\
\bottomrule
\end{tabular}
\caption{Conditional raw-prefix behavior on shared inputs. Insufficient reference hits fail the evidence-sensitive contract. Other prefix arms appear in Appendix~\ref{app:second-reader}.}
\label{tab:conditional}
\end{table}

Table~\ref{tab:conditional} localizes the failure. Qwen answers correctly on 79.8--83.7\% of partial sufficient inputs across arms, but abstains on only 1.0--5.0\% of insufficient inputs. Its raw-prefix concrete answers include 87 unsupported reference matches. DeepSeek concretely answers 182/203 insufficient raw inputs (89.66\%, CI $[84.78,94.02]$), with 85 reference hits (41.87\%, $[30.08,53.33]$) and 21 abstentions (10.34\%, $[5.91,15.05]$). Both prespecified replication thresholds are met. Across prefix arms, DeepSeek answers 71.2--74.0\% of partial sufficient cases correctly, while abstaining on only 9.9--14.0\% of insufficient cases. The diagnosis persists under explicit instructions against guessing.

First-route questions differ from the other operators because an early accepted route can fix the answer. In raw prefixes, first-route appropriateness is 40/79 certified inputs, versus 13/76 for count, 19/76 for last route, and 18/76 for state. These are descriptive operator breakdowns, not independent hypothesis tests. They show why replacing every truncated-context label with ``unanswerable'' would mischaracterize this task.

\subsection{Full Information and Program Controls}
With all fields present, Qwen's full CSV and raw agree with 188/320 and 166/320 references: $+6.88$ pp, CI $[2.50,11.25]$. DeepSeek scores 196/320 and 198/320: $-0.625$ pp, $[-5.00,3.44]$. Thus a positive full-source CSV contrast is not established in both configurations. This does not test a model-by-format interaction: differing interval significance does not establish a significant difference between effects. Syntax and token length change together, and cross-reader comparisons also include date and server changes.

Full raw agreement is lower than raw-prefix agreement in chronological confirmation (166 versus 181 matches). This does not make incomplete evidence preferable: many prefix matches lack sufficient evidence. Nor does it prove that all full-context errors arise from distraction. The additional records, input length, and reasoning burden change together. Separate evidence and answer measures prevent either reading from being smuggled into a single score.

Two deterministic controls clarify the task's difficulty. An END-only program abstains on every partial prefix and runs the executable query on full logs. It gets 203/307, 203/307, and 179/298 certified cases appropriate in the three partial arms, respectively, but fails every partial sufficient case. A cutoff-aware program also uses the public cutoff and first-route conditions. It is appropriate on all 307, 307, and 298 certified cases, using zero model calls. It shares the public parser and sufficient rules with the certifier, so this is a protocol consistency check rather than independent validation of a novel solver. It also establishes that the tested fixed grammar does not require an LLM or hybrid method to solve certified cases.

\section{RQ3: Evaluator Reliability}
\label{sec:judge}
\subsection{Constructed Held-Out Controls}
The judge study uses the 120 reserved histories from the outcome-masking generator. It selects one query per history, balanced across the four operators, and constructs 16 candidates per selected query. Ten scalar controls include correct values, incorrect values, conflicting assertions in both orders, and a previously declared extended answer format. Six evidence controls pair complete or masked logs with a correct value, incorrect value, or abstention. They require two Boolean outputs: reference agreement and context appropriateness. A joint success requires both axes to be correct. All labels follow the executable semantics and a finite, explicit rule that every asserted value must agree; a correct last assertion does not cancel an earlier contradiction.

This is evaluator calibration on constructed candidates, not an additional reader experiment. The hosted judge identifier is \texttt{deepseek-ai/DeepSeek-V4-Flash}. The main configuration uses the API setting \texttt{reasoning\_effort=low}, temperature 0, and output limits of 1,024 tokens for scalar cases and 2,048 for two-axis cases. A paired 24-history subset uses the weaker \texttt{none} configuration with limits of 32 and 64. Thus the comparison changes both reasoning setting and output budget; it is not an isolated causal test of reasoning effort. Additional independent HTTP repetitions assess a prespecified conflict stress case.

We compare a finite program, an extended program supporting the declared extra format, reference-only scoring, the model judge, and a program-first hybrid. The hybrid uses program results when available and otherwise the corresponding model receipt. Hybrid routing is simulated offline on retained receipts; it does not produce extra independent judgments. The extended program is fixed on development data before the reserved evaluation.

\begin{table}[t]
\centering\small
\setlength{\tabcolsep}{3pt}
\begin{tabular}{@{}lrrr@{}}
\toprule
Evaluator & Coverage & Joint correct & False accepts \\
\midrule
Finite program & 1,680 & 1,680 & 0/480 \\
Extended program & 1,920 & 1,920 & 0/480 \\
Reference only & 1,680 & 1,320 & 120/480 \\
LLM low & 1,918 & 1,908 & 0/480 \\
Program-first + low & 1,920 & 1,920 & 0/480 \\
\bottomrule
\end{tabular}
\caption{Judge confirmation over 1,920 constructed cases from 120 reserved histories. Coverage and correctness have denominator 1,920. False accepts refer to 480 scalar negative controls; they exclude the distinct two-axis support errors. Uncovered or invalid cases do not receive credit.}
\label{tab:judge}
\end{table}

\subsection{High Overall Accuracy Does Not Close the Audit}
The low configuration is jointly correct on 1,908/1,920 cases (99.38\%; history-clustered CI $[98.91,99.74]$). Its two-axis joint score is 708/720 (98.33\%), reference-axis score 710/720 (98.61\%), and support-axis score 716/720 (99.44\%). It falsely treats two concrete answers under decisive missing outcomes as context-appropriate; two other outputs truncate. Eleven of its 12 failures have \texttt{NONE} or \texttt{UNSET} references. The full prespecified reliability gate is not passed, despite the high aggregate score and zero false accepts on the separate scalar controls.

On the same 384 paired cases, low is correct on 382 and none on 343. The difference is $+10.16$ pp, CI $[7.81,13.02]$, resampling 24 histories. This supports improvement between these complete API configurations on the selected controls. It does not establish a general preference for one model family or a single-parameter mechanism. Relative to low across 120 histories, both the hybrid and extended program improve by $+0.62$ pp, $[0.26,1.09]$. The extended program uses zero API calls and matches the hybrid's observed 1,920/1,920 score. There is consequently no observed advantage of hybrid judging over that stronger simple baseline.

Perfect sample performance also has a finite-sample boundary. If error-bearing histories were independent identically distributed Bernoulli observations, zero errors in 120 histories would still allow a one-sided 95\% upper bound of $1-0.05^{1/120}=2.47$\% on the proportion of histories containing any error. The deterministic templates and shared task semantics further restrict transfer. We therefore use program-verified labels for the controlled reader results, rather than treating the judge as a validated substitute for human evidence assessment.

\section{Discussion}
\paragraph{What the diagnosis establishes.}
Both readers frequently supply concrete values when compatible histories force different answers; some match the original reference. Sufficient partial histories also require answers, preventing a blanket abstention policy from solving the diagnostic. The evidence-sensitive contract fails in two hosted configurations, without estimating prevalence in deployed memory systems.

\paragraph{What a format gain would mean.}
Compaction changes record coverage, the cutoff frontier, partial-tail certifiability, and readability. Qwen's full-source control shows that reference agreement can change even with fixed field coverage. Neither budgeted contrast establishes a positive appropriateness effect, and DeepSeek does not replicate the positive full-source contrast. These results do not measure retrieval quality, causal memory organization, or semantic--episodic module interactions; those require system-level interventions and baselines.

\paragraph{Consequences for evaluation practice.}
A history-QA report should state what source was available to the reader and what public assumptions make it sufficient. It should separate agreement on hidden references from appropriate behavior under missing information, retain unresolved certification cases, and report their coverage. Evaluators need controls that isolate distinct error axes: passing scalar incorrect-answer tests does not establish correct handling of support or abstention. In executable domains, a deterministic parser and executor is a necessary baseline. When these are inadequate for natural language, their certificates can still identify a bounded subset rather than assign unsupported labels to the rest.

\paragraph{Scope of the evidence.}
The analysis covers short synthetic device histories, fixed English questions, and two hosted reader families. Qwen confirmation uses new generator draws; DeepSeek replication reuses Qwen-exposed inputs. Neither supplies a second executable task or external-domain validation. The natural-trajectory pilot lacks validated evidence labels. Transfer to independent tasks, longer histories, and natural language remains open.

\section{Conclusion}
Constructive same-context witnesses and public cutoff rules separate reference agreement from evidence sufficiency. Qwen and DeepSeek both often match hidden references without sufficient evidence and seldom abstain on insufficient inputs. Budgeted CSV appropriateness effects remain inconclusive for both; Qwen's positive full-source contrast is not established for DeepSeek. A held-out judge audit retains errors despite high accuracy and is matched by a deterministic baseline. This bounded evaluation diagnosis calls for identifying the reader's evidence boundary and validating the measurement tool separately from answer matching.

\section*{Limitations}
\textbf{Restricted language and environment.} All principal evidence labels apply to explicit device operations, a fixed English query grammar, and declared public rules. The histories are short and synthetic. Natural conversations have ambiguity, implicit events, uncertain temporal expressions, and knowledge outside the supplied source. Our legal-world certificates do not resolve these issues, and these experiments do not measure long-horizon memory scaling.

\textbf{Limited model coverage.} Two hosted readers are tested with \texttt{none} reasoning and a 64-token output cap, not across their full capability range. DeepSeek reuses Qwen-exposed inputs with the original Qwen packing; identical bytes do not imply equal native-token budgets. Dates and server implementations differ, so cross-reader contrasts do not isolate weights. Temperature zero does not ensure determinism, and server weights cannot be pinned. Successful contents are sampled once; reader variability and model-family-wide failure rates are not estimated. Controlled reader labels are program-verified, independent of the separate DeepSeek judge audit.

\textbf{Conservative and shared certification.} The certifier can leave inputs unclassified, especially when a prefix ends inside a field. Coverage varies by representation. The main paired estimate is conditional on certification in both arms and does not identify the effect on all questions. Python and SQL implementations are checked separately, but originate from the same task specification. The cutoff-aware baseline shares its parser and sufficient rules with the certifier. Their agreement cannot exclude an error in the shared specification.

\textbf{Constructed interventions and controls.} Explicit \texttt{WITHHELD} outcomes, a complete-log marker, chronological guarantees, and rule explanations are unusually strong cues. Their distribution is designed for diagnosis. The masking arms differ in duplicate multiplicity for 47 queries. Format comparisons change syntax and token length together. The broader unordered-prefix witness search is post-collection, and its results are reported separately. Judge candidates use a finite known grammar and an artificial mix of errors; high accuracy is not an estimate on unconstrained generated answers.

\textbf{Statistical and validation boundary.} Four questions and multiple arms share each history, so they are not independent observations. Clustered intervals account for this pairing but not generator misspecification, hosted model drift, or selection into the certified subset. The chronological confirmation contains 80 histories and cannot rule out small effects. A confidence interval including zero does not prove equivalence. No human answer or evidence adjudication was performed. Executable truth supports the controlled semantics only. We make no claim of human-level judging, natural-domain transfer, a new memory method's superiority, or a validated gain on a public benchmark.

\section*{Ethical Considerations}
The controlled histories describe fictitious device identifiers and channels, and contain no collected personal conversations. External API requests contain synthetic log content and instructions. Credentials are excluded from retained public-facing artifacts. The natural-trajectory pilot is reported only as exploratory context in Appendix~\ref{app:pilot}; its source material would require a separate licensing and privacy check before redistribution.

An overconfident history reader could misrepresent actions taken or device state. Our metric prescribes abstention when evidence is insufficient, but deployment decisions require application-specific costs and safeguards. The program certificates should not be presented to users as proof about arbitrary language or real systems. An AI coding and writing assistant (OpenAI Codex) assisted protocol development, literature search, experimental software, analysis checks, and manuscript drafting. No model is listed as an author. Human authors are responsible for the methods, results, citations, and writing. No independent human answer or evidence adjudication was performed.

\clearpage
\appendix
\section{Public Rules and Certificate Checks}
\label{app:rules}
The chronological reader prompt states the following guarantees before the available log: reports form a prefix of a chronological log ordered by event time; distinct event identifiers have distinct event times; repeated reports describe the same event; original event count and log length are unknown; a prefix can end inside a record; and \texttt{[END OF COMPLETE LOG]} certifies that the complete original log is present. The query cutoff is inclusive. Report identifiers and report times do not determine event order. Rejected actions change nothing, accepted routes set the channel, and accepted disables set \texttt{DISABLED}.

The prompt explicitly requests an answer only when the visible information and these rules uniquely determine it, explains that a complete record after the cutoff can establish completeness of its earlier scope, and notes that a visible first accepted route can determine the first channel early. The response must be a JSON object with exactly one field, \texttt{answer}, without explanation. Counts are integers; first and last use \texttt{NONE} for no route; states use a channel, \texttt{DISABLED}, or \texttt{UNSET}; insufficiency requires \undet. Query strings are instantiated from four fixed templates:
\begin{quote}\small
As of [cutoff], how many distinct ACCEPTED route operations affected device [device]?\par
As of [cutoff], what was the channel of the first ACCEPTED route operation for device [device]?\par
As of [cutoff], what was the channel of the last ACCEPTED route operation for device [device]?\par
As of [cutoff], what was the state of device [device]?
\end{quote}

For a positive certificate, the parser consumes consecutive complete records and preserves the partial trailing string. A full marker closes the entire source. A complete record after $T$ closes the query scope because all subsequent records have event times no earlier than that record. Duplicate reports do not create new events. For the first-route rule, a visible accepted route at or before $T$ has no earlier unseen predecessor because the prefix begins at the start of the chronological log. The public replay determines its first qualifying channel.

For a negative certificate, the stored witness contains the alternative reports, the original and alternative answer, the mechanism, and the SHA-256 digest of the visible context. The audit validates both histories, deduplicates them, recomputes the answer with independent SQL and Python paths, and invokes the exact packer and tokenizer on the alternative source. A certificate is accepted only if the packed UTF-8 bytes agree and answers differ. Original event counts and probabilities are not constraints on alternative worlds. Candidate search failure leaves the input unclassified. This is essential for partial tails: a character already visible cannot be altered merely because the parser did not consume its full record.

\section{Additional Reader Breakdowns}
\label{app:reader}
Table~\ref{tab:operators} reports Qwen operator-level results from the chronological confirmation. Every reference denominator is 80. Appropriate denominators exclude the unclassified cases for that operator and arm. These values are descriptive decompositions of the main run, not additional experiments.

\begin{table}[h]
\centering\small
\begin{tabular}{@{}llrr@{}}
\toprule
Context & Operator & Reference & Appropriate \\
\midrule
Raw prefix & count & 40/80 & 13/76 \\
 & first & 63/80 & 40/79 \\
 & last & 43/80 & 19/76 \\
 & state & 35/80 & 18/76 \\
Whole-record & count & 40/80 & 13/76 \\
 & first & 63/80 & 40/79 \\
 & last & 45/80 & 20/76 \\
 & state & 41/80 & 16/76 \\
CSV prefix & count & 45/80 & 16/74 \\
 & first & 67/80 & 46/76 \\
 & last & 41/80 & 25/74 \\
 & state & 42/80 & 21/74 \\
\bottomrule
\end{tabular}
\caption{Qwen operator breakdown on chronological confirmation.}
\label{tab:operators}
\end{table}

The chronological development run uses 16 histories and 320 requests. Reference counts are 34/64 for raw prefixes, 35/64 for whole-record prefixes, 37/64 for CSV prefixes, 32/64 for full raw, and 40/64 for full CSV. Corresponding appropriate counts are 28/64, 28/64, 30/60, 32/64, and 40/64. The development gate passes, after which confirmation is generated from the unchanged design and its independent random stream. Development performance is not pooled with confirmation.

The unordered development grid has reference counts raw/whole/CSV of 35/34/34 at 256 tokens, 51/52/57 at 512, 96/90/106 at 1,024, and 112/112/126 at 2,048. Each cell has 160 query conditions. Full raw and full CSV score 112 and 126, respectively. The later appended-event witness search applies to the 1,440 non-full conditions in the primary grid and leaves the originally frozen labels intact. The 1,280-token supplementary comparison is selected using the 90th percentile of complete CSV lengths rounded upward to a multiple of 256 before its new reader calls. It yields raw/whole/CSV reference counts of 108/107/126 and appropriate counts of 0/0/126 under the broader witness audit. No answer abstains. These are development observations under a different task from chronological confirmation.

\section{Second-Reader Replication Details}
\label{app:second-reader}
The replication reuses the 80 chronological confirmation histories and all 1,600 frozen visible inputs after Qwen results have been observed. No DeepSeek reader outputs on these inputs precede its freeze. The task, packers, seed, messages, output schema, temperature, reasoning setting, and 64-token cap are unchanged; only the model identifier changes. The eight technical requests are included in the main cohort. They check identity, service, output, and cost contracts without an answer-accuracy gate. All 1,600 responses finish normally and satisfy the answer contract. Reader answers are scored against executable labels, not by the DeepSeek judge.

Table~\ref{tab:replication-conditional} retains the conditional counts for all three prefix arms. Sufficient-case abstentions are errors; insufficient-case abstentions are appropriate. Unclassified cases are retained in overall reference denominators and excluded from certified appropriateness.

\begin{table}[t]
\centering\small
\setlength{\tabcolsep}{3pt}
\begin{tabular}{@{}llrrr@{}}
\toprule
Reader & Certified subset & Raw & Whole & CSV \\
\midrule
Qwen & S: correct & 83/104 & 87/104 & 99/119 \\
 & S: abstain & 4/104 & 0/104 & 1/119 \\
 & I: concrete & 196/203 & 201/203 & 170/179 \\
 & I: abstain & 7/203 & 2/203 & 9/179 \\
 & I: ref. hit & 87/203 & 91/203 & 75/179 \\
\midrule
DeepSeek & S: correct & 77/104 & 74/104 & 86/119 \\
 & S: abstain & 23/104 & 21/104 & 20/119 \\
 & I: concrete & 182/203 & 183/203 & 154/179 \\
 & I: abstain & 21/203 & 20/203 & 25/179 \\
 & I: ref. hit & 85/203 & 88/203 & 62/179 \\
\bottomrule
\end{tabular}
\caption{Both readers on the same certified subsets. S and I denote partial sufficient and certified insufficient.}
\label{tab:replication-conditional}
\end{table}

The replication criterion applies to the 203 insufficient raw-prefix cases: the lower 95\% endpoint must exceed 50\% for concrete answering and zero for unsupported reference agreement. Both pass. These thresholds diagnose this fixed configuration and do not estimate a universal failure rate. The whole-record and CSV diagnostics retain their clustered intervals in the recorded summary. No sample size or setting is adapted after seeing outputs.

Table~\ref{tab:replication-contrasts} reports every declared within-DeepSeek contrast. CSV versus raw appropriateness uses 285 jointly certified questions, omitting 35 unclassified on at least one side. Whole versus raw appropriateness uses 307. Reference contrasts use all 320 queries. All intervals resample 80 histories, not individual queries, with 10,000 replicates. Intervals are reported individually without simultaneous error control.

\begin{table}[t]
\centering\small
\setlength{\tabcolsep}{3pt}
\begin{tabular}{@{}lrrl@{}}
\toprule
DeepSeek contrast & $n$ & $\Delta$ pp & 95\% CI \\
\midrule
CSV--raw, appropriate & 285 & $+2.46$ & $[-1.84,6.94]$ \\
CSV--raw, reference & 320 & $-1.56$ & $[-5.94,2.81]$ \\
Whole--raw, appropriate & 307 & $-1.30$ & $[-4.82,2.66]$ \\
Full CSV--raw, reference & 320 & $-0.62$ & $[-5.00,3.44]$ \\
\bottomrule
\end{tabular}
\caption{Prespecified secondary contrasts for the second reader. $n$ counts paired queries, not independent histories. The full-source point estimate is exactly $-0.625$ pp.}
\label{tab:replication-contrasts}
\end{table}

Table~\ref{tab:reader-config} gives all prespecified cross-configuration comparisons. Because hosted execution dates and server implementations differ, these are descriptive comparisons rather than isolated effects of model weights. They do not test model-by-format interactions. Full-source reference and appropriate indicators coincide, but their separately seeded bootstrap calculations can differ slightly in endpoints.

\begin{table*}[t]
\centering\small
\begin{tabular}{@{}lrrlrrl@{}}
\toprule
 & \multicolumn{3}{c}{Reference agreement} & \multicolumn{3}{c}{Context appropriateness} \\
\cmidrule(lr){2-4}\cmidrule(l){5-7}
Context & $n$ & $\Delta$ pp & 95\% CI & $n$ & $\Delta$ pp & 95\% CI \\
\midrule
Raw prefix & 320 & $-2.81$ & $[-8.12,2.19]$ & 307 & $+2.61$ & $[-2.73,7.92]$ \\
Whole-record raw & 320 & $-5.31$ & $[-10.00,-0.94]$ & 307 & $+1.63$ & $[-3.90,7.12]$ \\
CSV prefix & 320 & $-8.75$ & $[-13.75,-4.06]$ & 298 & $+1.01$ & $[-4.91,7.02]$ \\
Full raw & 320 & $+10.00$ & $[1.25,18.44]$ & 320 & $+10.00$ & $[1.56,18.44]$ \\
Full CSV & 320 & $+2.50$ & $[-5.31,10.62]$ & 320 & $+2.50$ & $[-5.62,10.31]$ \\
\bottomrule
\end{tabular}
\caption{DeepSeek minus historical Qwen on shared texts; 80-history paired bootstrap. Date and service changes are confounded with reader identity.}
\label{tab:reader-config}
\end{table*}

\section{Judge Controls and Failure Localization}
\label{app:judge}
The judge confirmation balances one selected query per history across count, first, last, and state, 30 histories each. Fifteen histories with no accepted target event are distributed 4/4/4/3 across these operators. The weaker-configuration paired subset selects six histories per operator before requests. Eight original scalar templates cover four positive renderings, two wrong values, and contradictory value orderings. Two further templates use the declared extended format ``Final answer for the requested device: VALUE.'' The finite program declines this extra syntax; the frozen extended program handles it.

The six evidence controls and the low judge's joint correctness are: complete/correct 120/120; complete/wrong 118/120; decisive mask/reference value 118/120; decisive mask/abstention 114/120; complete/abstention 118/120; irrelevant mask/correct 120/120. Together these yield 708/720 joint successes. Two decisive-mask/reference cases are falsely supported. Two outputs truncate. The scalar negative-control false-accept count is zero, which cannot be substituted for the two-axis support-error count.

The full gate combines overall accuracy, scalar positive and negative error rates, both evidence axes, joint accuracy, abstention behavior, extended-format performance, coverage, contract validity, and a three-call conflict stress test. Its thresholds are operational sample thresholds, not population guarantees. In the paired stress subset, none has one changing group among 24 histories and 15 stably wrong groups. Repeated HTTP calls are counted as repetitions of their original history, not new independent examples.

Earlier development compares additional reasoning settings and hybrid routing, but these are not independent confirmation. An extended program rule, introduced on development examples, covers the entire declared candidate grammar; it is already fixed before the reserved study. In a supplementary offline evaluation of all four query types on the same 120 histories, it scores 7,680/7,680, versus the finite program's 6,720 covered cases. This is a grammar-coverage check, not evidence for general semantic evaluation.

\section{Exploratory Natural-Trajectory Pilot}
\label{app:pilot}
An earlier pilot uses six exposed AMA episodes, 18 questions, and 198 answers over 11 memory/context conditions. It motivates the subsequent evaluator audit but is not part of the controlled confirmation. A separate blind judge yields 193 valid judgments out of 198 and disagrees with the original judge on 59 of those 193. Its calibration is correct on 101/108 constructed controls, with five false accepts among 54 negative controls and two invalid outputs, failing the fixed reliability gate. Forty-four candidate outputs are preflagged for reference or source risks. No human adjudication resolves them.

The pilot neither estimates full AMA-Bench performance nor validates an architecture improvement. It also contains packaging differences between arms and no count or latest-state queries. We therefore do not use its apparent memory-condition ranking as evidence for the main claims. It does not provide an external validation set for the executable evidence certificates. Reporting this unsuccessful validation path prevents exploratory natural-data judgments from being mistaken for audited ground truth.

\section{Reproducibility and Resource Accounting}
\label{app:reproduce}
The local research record retains generation and scoring scripts, public prompts, ground-truth files, both-world witnesses, input/request hashes, sanitized raw receipts, fee estimates, and offline replay checks. It also records which histories were exposed during development. These materials have not been released as an anonymous public artifact. The intended release comprises original experiment code and program-generated logs, excluding pretrained model weights and AMA-Bench source histories.

\paragraph{Artifact sources and software.}
Our retained Qwen3-32B \texttt{tokenizer.json} matches the official repository file,\footnote{\url{https://huggingface.co/Qwen/Qwen3-32B/blob/c6740cde69fed4893acae7deae8280ce459e827f/tokenizer.json}} with SHA-256 \texttt{aeb13307\allowbreak a71acd8f\allowbreak e81861d9\allowbreak 4ad54ab6\allowbreak 89df7733\allowbreak 18809eed\allowbreak 3cbe794b\allowbreak 4492dae4}. We load it using Hugging Face Tokenizers \citep{moi2023tokenizers}, with \texttt{add\_special\_tokens=False}; no tokenizer is trained. The initial preflight records Tokenizers 0.21.1; later stages freeze the tokenizer file but do not independently record the library version.

Frozen main-run records list Python 3.9.13 from the Python Software Foundation,\footnote{\url{https://docs.python.org/3.9/}} NumPy 1.26.4 \citep{harris2020numpy} for numerical analysis, and Requests 2.32.3 by Kenneth Reitz and contributors\footnote{\url{https://requests.readthedocs.io/}} for HTTP calls. SQL checks use Python's \texttt{sqlite3} interface to SQLite, credited to the SQLite project.\footnote{\url{https://www.sqlite.org/}} Matplotlib \citep{hunter2007matplotlib} generates local diagnostic plots. Historical Matplotlib and SQLite engine versions are not recorded. Scoring, certification, and bootstrap logic are implemented in our experiment scripts. Model citations identify the advertised model families; they do not independently establish the hosted weight snapshots.

The chronological design seed is 2026100716. Confirmation uses the fixed independent offset 1000003 and history increment 7919 in its pseudorandom stream. Its 80 confirmation histories contain four queries in five contexts, giving 1,600 unique requests; the 16-history development adds 320. Four confirmation requests are interrupted locally without a recoverable response or usage record. The audit cannot determine whether the server processed or billed them. They are retried once with the same payload and a conservative maximum-cost reserve; unreceived contents are not scored. Across both Qwen stages there are 1,920 successful unique requests and 1,924 transport attempts. Known API fee estimates total USD 0.29368444, with USD 0.00064232 reserved for unknown outcomes, giving USD 0.29432676 accounted against a USD 0.65 cap.

The DeepSeek replication uses the unchanged payload seed 2026100716 and request-order seed 2026100917. Its 1,600 unique requests require 1,600 transport attempts, with no failed transports, unknown-cost reserves, or content resampling. Estimated API cost is USD 0.10184188 against a USD 0.65 cap. Recorded native API usage is 2,310,812 prompt tokens (1,507,072 cached) and 13,211 completion tokens, with zero observed reasoning tokens. Frozen prices are USD 0.09 per million input tokens, 0.018 per million cached input tokens, and 0.18 per million output tokens. Every receipt cost is recomputed. The audit reexecutes all 585 insufficiency witnesses, verifies model-only payload changes and 1,600 unique response identifiers disjoint from Qwen, and confirms that prior frozen results are unchanged. Network-disabled replay reproduces scored rows and the summary byte-for-byte.

The judge confirmation uses seed 2026100714 and 2,400 requests, including 1,920 main low cases, 384 paired none cases, and 96 extra stress calls. All receipts have independent nonempty response identifiers. Estimated API cost is USD 0.26457120 under a USD 0.70 cap. Hosted prices are frozen at execution and fees are recomputed from recorded input, cache, and output usage; estimates do not imply a verified external invoice. The reader and judge are accessed through DeepInfra's OpenAI-compatible hosted endpoint,\footnote{\url{https://deepinfra.com/}} with literal model identifiers retained in the manifests. The server weights cannot be independently pinned.

Retries concern documented transport failures, not content selection. Statistical replay disables the request library's network access and reproduces retained summary outputs byte-for-byte. This checks reconstruction of the recorded results; it does not claim that a fresh API run would return identical contents.

\end{document}